\documentclass[twoside,11pt]{article}
\usepackage{jmlr2e}

\ShortHeadings{Variable Importance Scores}{Loh and Zhou}
\firstpageno{1}

\usepackage{graphics}
\usepackage{pstricks,pst-node,pst-tree}
\usepackage{longtable}
\usepackage{lscape}

\usepackage{url}
\usepackage{breakurl}
\usepackage{ragged2e}

\begin{document}

\title{Variable Importance Scores}

\author{\name Wei-Yin Loh \email loh@stat.wisc.edu \\
       \addr Department of Statistics\\
       University of Wisconsin\\
       Madison, WI 53706, USA
       \AND
       \name Peigen Zhou \email pzhou9@wisc.edu \\
       \addr Department of Statistics\\
       University of Wisconsin\\
       Madison, WI 53706, USA}

\editor{}

\maketitle

\begin{abstract}
  Scoring of variables for importance in predicting a response is an
  ill-defined concept. Several methods have been proposed but little
  is known of their performance.  This paper fills the gap with a
  comparative evaluation of eleven methods and an updated one based on
  the GUIDE algorithm.  For data without missing values, eight of the
  methods are shown to be biased in that they give higher or lower
  scores to different types of variables, even when all are
  independent of the response. Of the remaining four methods, only two
  are applicable to data with missing values, with GUIDE the only
  unbiased one. GUIDE achieves unbiasedness by using a
  self-calibrating step that is applicable to other methods for score
  de-biasing.  GUIDE also yields a threshold for distinguishing
  important from unimportant variables at 95 and 99 percent confidence
  levels; the technique is applicable to other methods as well.
  Finally, the paper studies the relationship of the scores to
  predictive power in three data sets.  It is found that the scores of
  many methods are more consistent with marginal predictive power than
  conditional predictive power.
\end{abstract}


\section{Introduction}
\label{sec:intro}
The question of how to quantify the relative importance of variables
has intrigued researchers for years.  While it was largely of academic
interest early on, the question has taken on greater urgency in the
last two decades, due to the increasing frequency of large data sets
and the popularity of ``black box'' machine learning methods for which
scoring the importance of variables may be the only means of
interpretation; see \citet{Bring94}, \citet{Bi12} and \citet{Wei15}
for excellent surveys.  A prime black-box example is random forest
(RF, \citet{RF}), which consists of hundreds of unpruned regression
trees.  Its permutation-based scheme to produce importance scores has
been copied by many methods.

Some researchers have observed that the orderings of RF scores do not
always agree with those based on traditional methods.
\citet{Bureau05} used RF to identify single-nucleotide polymorphisms
(SNPs) predictive of disease and found that while SNPs that are highly
associated with disease, as measured by Fisher's exact test, tend to
have high RF scores, the two orderings do not match.  \citet{Diaz06}
selected genes in microarray data by iteratively removing 20\% of the
genes with the lowest RF scores at each step.  They found that this
yielded a smaller set of genes than linear discriminant analysis,
nearest neighbor and support vector machine methods, and that the RF
results were more variable.

\begin{table}
  \centering
  \caption{Variables in COVID data}
  \label{tab:covid:vars} \vspace{0.5em}
  \begin{tabular}{lp{4.7in}} \hline
    \texttt{died} & Died while hospitalized (0=no, 1=yes) \\
    \texttt{agecat} & Age group (0=18--50, 1=50--59, 2=60--69,
                      3=70--79, 4=80--90 years) \\
    \texttt{race} & White; Black or African American; Asian; Native
                    Hawaiian or other Pacific Islander; American Indian or
                    Alaska Native; Unknown \\
    \texttt{sex} & Gender (male/female) \\
    \texttt{aids} & AIDS/HIV (0=no, 1=yes) \\
    \texttt{cancer} & Any malignancy, including lymphoma and leukemia,
                      except malignant neoplasm of skin (0=no, 1=yes) \\ 
    \texttt{cerebro} & Cerebrovascular disease (0=no, 1=yes) \\
    \texttt{charlson} & Charlson comorbidity index (0--20) \\ 
    \texttt{CHF} & Congestive heart failure (0=no, 1=yes) \\
    \texttt{CPD} & Chronic pulmonary disease (0=no, 1=yes) \\
    \texttt{dementia} & Dementia (0=no, 1=yes) \\
    \texttt{diabetes} & Diabetes mellitus (0=no, 1=yes) \\
    \texttt{hemipara} & Hemiplegia or paraplegia (0=no, 1=yes) \\
    \texttt{metastatic} & Metastatic solid tumor (0=no, 1=yes) \\
    \texttt{MI} & Myocardial infarction (0=no, 1=yes) \\
    \texttt{mildliver} & Mild liver disease (0=no, 1=yes) \\
    \texttt{modsevliv} & Moderate/severe liver disease (0=no, 1=yes) \\
    \texttt{PUD} & Peptic ulcer disease (0=no, 1=yes) \\
    \texttt{PVD} & Peripheral vascular disease (0=no, 1=yes) \\
    \texttt{RD} & Rheumatic disease (0=no, 1=yes) \\
    \texttt{renal} & Renal disease (0=no, 1=yes) \\
    \hline
  \end{tabular}
\end{table}



The differences in orderings may be demonstrated on a data set from
\citet{Harrison20} of 31,461 patients aged 18--90 years diagnosed with
the COVID-19 disease between January 20 and May 26, 2020, in the
United States.  Table~\ref{tab:covid:vars} lists the 21 variables,
which consist of death during hospitalization, age group, sex, race,
16 comorbidities, and Charlson comorbidity index (a risk score
computed from the comorbidities).  The authors estimated mortality
risk by fitting a multiple linear logistic regression model, without
Charlson index, to each age group. They found 10 variables
statistically significant at the 0.05 level (without multiplicity
adjustment), namely, race, sex, and history of myocardial infarction
(\texttt{MI}), congestive heart failure (\texttt{CHF}), dementia,
chronic pulmonary disease (\texttt{CPD}), mild liver disease
(\texttt{mildliver}), moderate/severe liver disease
(\texttt{modsevliv}), renal disease (\texttt{renal}), and metastatic
solid tumor (\texttt{metastatic}).

\begin{figure}
  \centering
  \resizebox{0.95\textwidth}{!}{\includegraphics*[46,25][586,771]{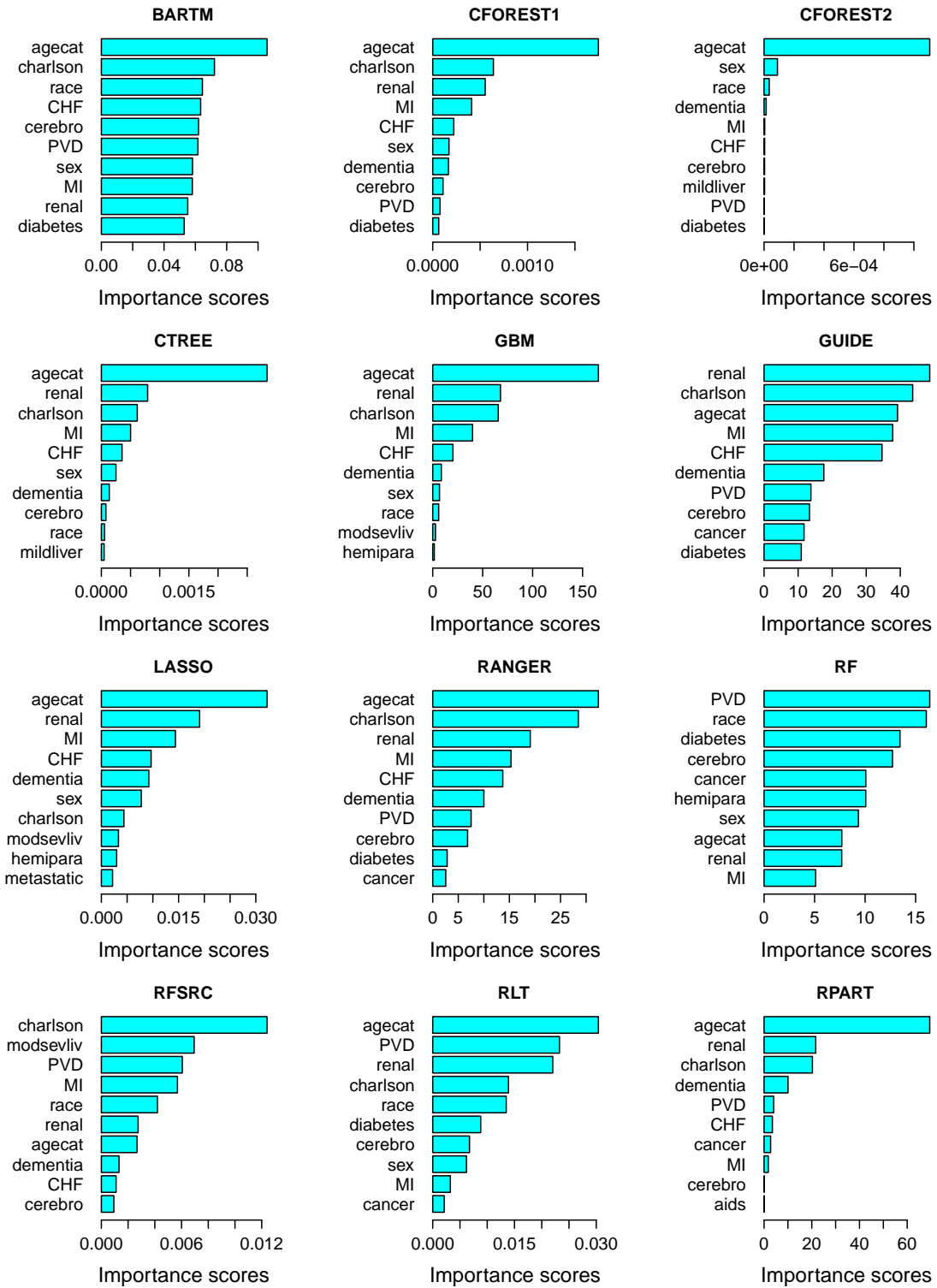}}
  \caption{Top 10 variables for COVID data; scores of LASSO, RANGER,
    RF, RFSRC, and RLT are averaged over 100 runs with different
    random seeds}
  \label{fig:covid:barplots}
\end{figure}

Figure~\ref{fig:covid:barplots} shows the importance scores of the top
10 variables obtained from 12 methods discussed below.  There is
substantial variation in the orderings, although \texttt{agecat},
\texttt{charlson}, and \texttt{renal} are ranked in the top 3 by 7 of
the 12 methods. Of the variables that \citet{Harrison20} found
statistically significant, \texttt{CPD} is not ranked in the top 10 by
any method, and \texttt{mildliver} and \texttt{metastatic} are ranked
in the top 10 only twice and once, respectively.  On the other hand,
the non-significant variables \texttt{cancer}, \texttt{cerebro},
\texttt{diabetes}, \texttt{hemipara}, and \texttt{PVD} are ranked in
the top 10 by 5, 10, 7, 3, and 9 methods, respectively.  Statistical
significance is clearly not necessarily consistent with the importance
scores.

What is one to do in the face of such disparate results? One solution
is to average the ranks across the methods, but this assumes that the
methods are equally good.  \citet{Strobl07}, \citet{SZ08}, and others
have shown that the scores from RF are unreliable because they are
biased towards certain variable types. A method is said to be
``unbiased'' if all predictor variables have the same mean importance
score when they are independent of the response variable.  One goal of
this paper is to find out if other methods are biased.

\begin{figure}
  \centering
  \resizebox{0.9\textwidth}{!}{\includegraphics*[17,29][576,763]{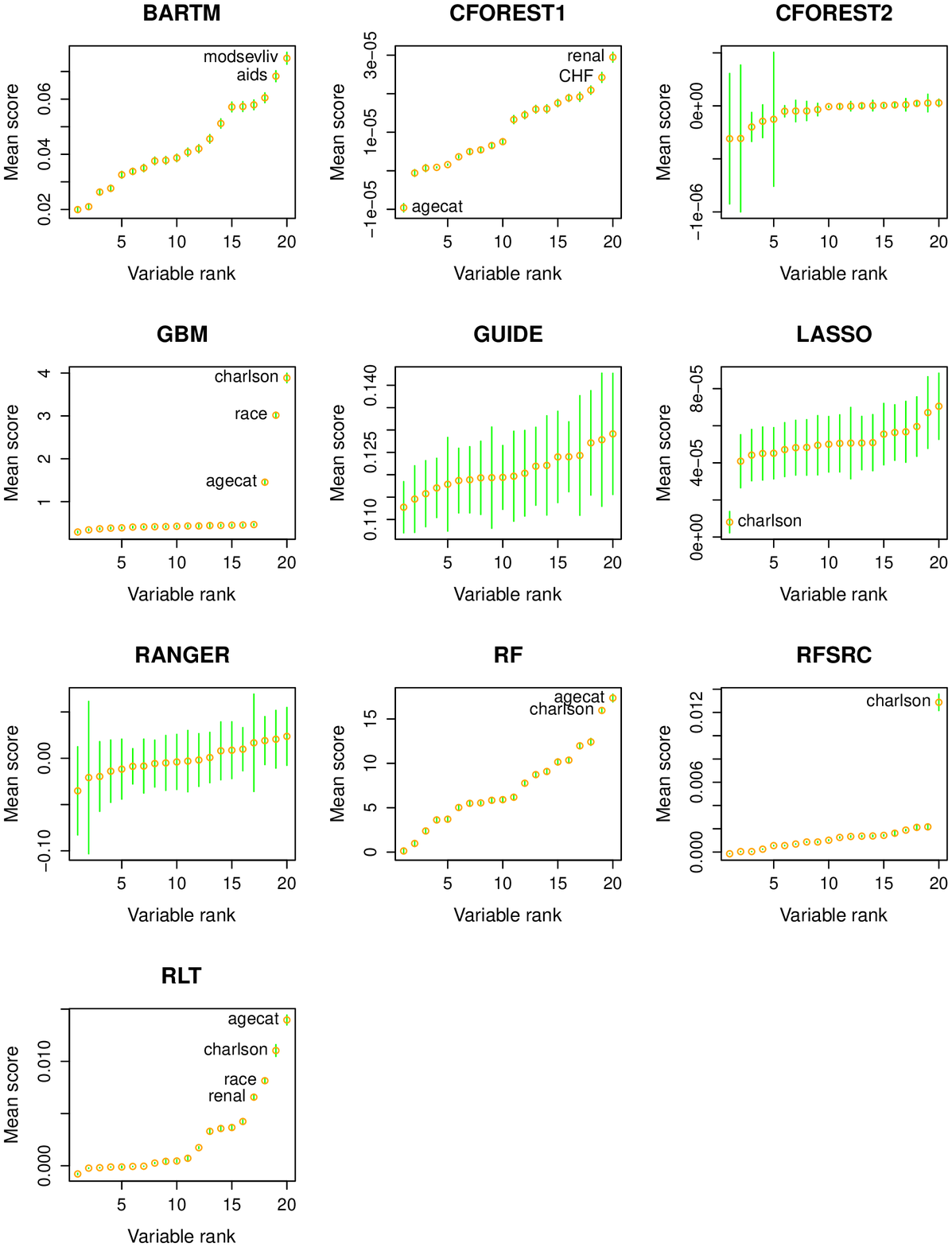}}
  \caption{Mean importance scores $\overline{\mathtt{VI}}$ (orange)
    and 2-SE bars (green) from 1000 random permutations of the
    dependent variable for COVID data. Variables ordered by increasing
    mean scores.  CTREE and RPART are not included because they
    returned trees with no splits (and hence no importance scores) for
    all permutations.}
  \label{fig:covid:perm}
\end{figure}

Given a data set, bias may be uncovered by estimating the mean scores
over random permutations of the response variable, keeping the values
of the predictor variables fixed.  Let $\mathtt{VI}_j(X)$
($j=1,2,\ldots,J$) denote the importance score of variable $X$ in the
$j$th permutation.  Figure~\ref{fig:covid:perm} plots the values of
$\overline{\mathtt{VI}}(X) = J^{-1} \sum_j \mathtt{VI}_j(X)$ in
increasing order and their 2-standard error bars, for $J=1000$.  An
unbiased method should have all its error bars overlapping.  The plots
show that only CFOREST2, GUIDE, and RANGER have this property.

Another interesting problem is to identify variables that are truly
important.  There few attempts at answering this question, despite its
being central to variable selection---an important step if the total
number of variables exceeds the sample size. Although it may be
expected that all important variables should be included,
\citet{lchen12} showed that under certain conditions, omitting the
less important ones can yield a model with higher prediction accuracy.

The remainder of this article is organized as follows.
Section~\ref{sec:guide} describes the GUIDE method of calculating
importance scores.  Section~\ref{sec:methods} reviews the other 11
methods.  Section~\ref{sec:simul} presents the results of simulations
to identify the biased methods and show their effects on the scores.
Section~\ref{sec:predict} examines the extent to which the scores of
each method are consistent with two measures of predictive power of
the variables. Section~\ref{sec:threshold} describes a general
procedure for producing a threshold score such that, with high
probability, variables with scores less than the threshold are
independent of the response.  Section~\ref{sec:miss} shows how the
GUIDE method applies to data with missing values.
Section~\ref{sec:conclusion} concludes the article with some remarks.

\section{GUIDE}
\label{sec:guide}
The GUIDE algorithm for constructing regression and classification
trees is described in \citet{guide} and \citet{guide09}, respectively.
It differs from CART \citep{cart} in every respect except tree
pruning, where both employ the same cost-complexity cross-validation
technique.  CART uses greedy search to select the split that most
decreases node impurity but GUIDE uses chi-squared tests to first
select a split variable and then searches for the best split based on
it that most decreases node impurity. This approach started in
\citet{lv88} and evolved principally through \citet{chly94},
\citet{lohshih97}, and \citet{guide,guide09}. Besides reducing
computation, it lets GUIDE avoid biases in variable selection inherent
among greedy search methods. Another major difference between GUIDE
and CART is how each deals with missing values in predictor
variables. \citet{cruise} showed that CART's solution through
surrogate splits is another source of selection bias.

An initial importance scoring method based on GUIDE was proposed in
\citet{lchen12}. Though not designed to be unbiased, it turned out to
be approximately unbiased unless there is a mix of ordinal and
categorical variables. We present here an improved version for
regression that ensures unbiasedness. As in the previous method, it
uses a weighted sum of chi-squared statistics obtained from a shallow
(four-level) unpruned tree, but it adds conditional tests for
interaction and a permutation-based step for bias adjustment.  Given a
node $t$, let $n_t$ denote the number of observations in $t$.
\begin{enumerate}
\item Fit a constant to the data in $t$ and compute the residuals.
\item Define a class variable $Z$ such that $Z=1$ if the observation
  has a positive residual and $Z=2$ otherwise.
\item 
  \begin{enumerate}
  \item If $X_k$ is an ordinal variable, transform it to a categorical
    variable $X_k'$ with $m$ roughly equal-sized categories, where
    $m=3$ if $n_t < 60$ and $m=4$ otherwise.
  \item If $X_k$ is a categorical variable, define $X_k' = X_k$.
  \end{enumerate}
\item If $X_k$ has missing values, add an extra category to $X_k'$ to
  hold the missing values.
\item For $k=1,2, \ldots, K$, where $K$ is the number of variables,
  perform a contingency table chi-squared test of $X_k'$ versus $Z$
  and denote its p-value by $p_1(k,t)$.
\item \label{step:7} If $\min_k p_1(k,t) \geq 0.10/K$ (first
  Bonferroni correction), carry out the following interaction tests.
  \begin{enumerate}
  \item Transform each ordinal $X_k$ to a 3-level categorical variable
    $X_k'$.  If $X_k$ has no missing values, $X_k'$ is $X_k$
    discretized at the 33rd and 67th sample quantiles.  If $X_k$ has
    missing values, $X_k'$ is $X_k$ discretized at the sample median
    with missing values forming the third category.  If $X_k$ is a
    categorical variable, let $X_k' = X_k$.
  \item For every pair $(X'_j, X'_k)$ with $j < k$, perform a
    chi-squared test with the $Z$ values as rows and the
    $(X'_j, X'_k)$ values as columns and let $p_2(j,k,t)$ denote its
    p-value.
  \item Let $(X_{j'}', X_{k'}')$ be the pair of variables with the
    smallest value of $p_2(j,k,t)$.  If
    $p_2(j',k',t) < 0.20\{K(K-1)\}^{-1}$ (second Bonferroni
    correction), redefine $p_1(j',t) = p_1(k',t) = p_2(j',k',t)$.
  \end{enumerate}
\item Let $k^*$ be the smallest value of $k$ such that
  $p_1(k^*,t) = \min_k p_1(k,t)$. Find the split on $X_{k^*}$ yielding
  the largest decrease in node impurity (i.e., sum of squared residuals).
\end{enumerate}
After a tree is grown with four levels of splits, the importance score
of $X_k$ is computed as
\begin{equation}
  \label{eq:guide:raw}
  v(X_k) = \sum_t \sqrt{n_t} \, \chi^2_1(k,t)
\end{equation}
where the sum is over the intermediate nodes and $\chi^2_1(k,t)$
denotes the $(1-p_1(k,t))$-quantile of the chi-squared distribution
with 1 degree of freedom. The factor $\sqrt{n_t}$ in
(\ref{eq:guide:raw}) first appeared in \citet{lchen12} but was changed
to $n_t$ in \citet{lhm15}; we revert it back to $\sqrt{n_t}$ to
prevent the root node from dominating the scores.

The values of $v(X_k)$ are slightly biased due partly to differences
between ordinal and categorical variables and partly to the above
step~\ref{step:7}. To remove the bias, we adjust the scores by their
means computed under the hypothesis that the response variable ($Y$)
is independent of the $X$ variables. Specifically, the $Y$ values are
randomly permuted $B$ times (the default is $B=300$) with the $X$
values fixed, and a tree with four levels of splits is constructed for
each permuted data set.  Let $v^*_b(X_k)$ be the value
of~(\ref{eq:guide:raw}) in permutation $b=1,2,\ldots,B$, and define
$\bar{v}(X_k) = B^{-1} \sum_b v^*_b(X_k)$.  The GUIDE
\emph{bias-adjusted} variable importance score of $X_k$ is
\begin{equation}
  \label{eq:guide:scaled}
  \mathtt{VI}(X_k) = v(X_k) / \bar{v}(X_k).
\end{equation}


\section{Other methods}
\label{sec:methods}
We briefly review the other methods here.
\begin{description}
\item[RPART.] This is an R version of CART \citep{rpart}.  Let
  $s = \{X_i \in A\}$ denote a split of node $t$ for some variable
  $X_i$ and set $A$, and let $t_L$ and $t_R$ denote its left and right
  child nodes.  Given a node impurity function $i(t)$ at $t$, let
  $\Delta(s,t) = i(t)-i(t_L)-i(t_R)$ be a measure of the goodness of
  the split.  For regression trees,
  $i(t) = \sum_{i \in t} (y_i-\bar{y}_t)^2$, where $\bar{y}_t$ is the
  sample mean at $t$. CART partitions the data with the split $s(t)$
  that maximizes $\Delta(s,t)$.  To evaluate the importance of the
  variables as well as to deal with missing values, CART finds, for
  each $X_j$ ($j \neq i$), the surrogate split $\tilde{s}_j(t)$ that
  best predicts $s(t)$.  The importance score of $X_j$ is
  $\sum_t \Delta(\tilde{s}_j(t),t)$, where the sum is over the
  intermediate nodes of the pruned tree \citep[p.~141]{cart}.

  RPART measures importance differently from CART
  \citep{rpartvignette}.  Given a split $s(t)$ and a surrogate
  $\tilde{s}(t)$, let $k(s(t), \tilde{s}(t))$ be the total number of
  observations in $t_L$ and $t_R$ correctly sent by $\tilde{s}(t)$.
  Let $n_L$ and $n_R$ denote the number of observations in $t_L$ and
  $t_R$, respectively.  The ``adjusted agreement'' between $s$ and
  $\tilde{s}$ is
  $a(s,\tilde{s}) = \{k(s,\tilde{s})-\max(n_L,n_R)\}/\min(n_L,n_R)$.
  Call $X_i$ a ``primary'' variable if it is in $s$ and a
  ``surrogate'' variable if it is in $\tilde{s}$.  Let $P(i)$ and
  $S(i)$ denote the sets of intermediate nodes where $X_i$ is the
  primary and surrogate variable, respectively.  RPART defines
  $\mathtt{VI}(X_i) = \sum_{t \in P(i)} \Delta(s(t),t) + \sum_{t \in
    S(i)} a(s(t),\tilde{s}(t))\Delta(\tilde{s}(t),t)$
  as the importance score of $X_i$. As shown below, this method yields
  biased scores, because maximizing the decrease in node impurity
  induces a bias towards selecting variables that allow more splits
  \citep{WL94,lohshih97} and the surrogate split method itself induces
  a bias when there are missing values \citep{cruise}. 
\item[GBM.]  This is gradient boosting machine \citep{Friedman01}. It
  uses functional gradient descent to build an ensemble of short CART
  trees.  For a single tree, the importance score of a variable is the
  square root of the total decrease in node impurity (squared error in
  the case of regression) over the nodes where the variable appears in
  the split. For an ensemble, it is the root mean squared importance
  score of the variable over the trees \citep[p.~1217]{Friedman01}.
  We use the R function \texttt{gbm} \citep{gbm-man} to construct the
  GBM models and the \texttt{varImp} function in the \texttt{caret}
  package \citep{caret} to calculate the importance scores.

\item[RF.] This is the R implementation of random forest
  \citep{Liaw02}.  It has two measures for computing importance
  scores. The first is the ``decrease in accuracy'' of the forest in
  predicting the ``out-of-bag'' (OOB) data before and after random
  permutation of the predictor variable, where the OOB data are the
  observations not in the bootstrap sample. The second uses the
  ``decrease in node impurity,'' which is the average of the total
  decrease in node impurity of the trees.  Partly due to CART's split
  selection bias, the decrease in node impurity measure is known to be
  unreliable \citep{Strobl07,SZ08}.  The results reported here use the
  ``decrease in accuracy'' measure.

\item[RANGER.]  \citet{SZ08} used pseudovariables to correct the bias
  in RF's ``decrease in node impurity'' method. (Pseudovariables were
  employed earlier by \citet{WBS07}.)  Given $K$ predictor variables
  $\mathbf{X} = (X_1, X_2, \ldots, X_K)$, another $K$ pseudovariables
  $\mathbf{Z} = (Z_1, Z_2, \ldots, Z_K)$ are added where the rows of
  $\mathbf{Z}$ are random permutations of the rows of
  $\mathbf{X}$. The RF algorithm is applied to the $2K$ predictors and
  the importance score of $X_i$ is adjusted by subtracting the score
  of $Z_i$ for $i=1,2,\ldots,K$.  This approach requires more computer
  memory and increases computation time (a forest has to be
  constructed for each generation of $\mathbf{Z}$).  \citet{NKW18}
  proposed using only a single generation of $\mathbf{Z}$ and storing
  only the permutation indices rather than the values of $\mathbf{Z}$.
  Their method is implemented in the \texttt{ranger} R package
  \citep{ranger}. Although storing only the permutation indices saves
  computer memory, the use of a single permutation adds another level
  of randomness to the already random results of RF.  In serious
  applications, there are no savings in computation time because
  RANGER must be applied many times to stabilize the average
  importance scores. In the real data examples here, the RANGER scores
  are averages over 100 replications.

\item[RFSRC.]  This is another ensemble method similar to RF
  \citep{Ishwaran07,Ishwaran08}.  The importance of a variable $X$ is
  measured by the difference between the prediction error of the OOB
  sample before and after $X$ is ``noised up''. ``Noising up'' here
  means that if an OOB observation encounters a split on $X$ at a node
  $t$, it is randomly sent to the left or right branch, with equal
  probability, at $t$ and \emph{all} its descendent nodes.  Missing
  values in a predictor variable are imputed nodewise, by replacing
  each missing value with a randomly selected non-missing value in the
  node. The results for RFSRC here are obtained with the
  \texttt{randomForestSRC} R package \citep{rfsrc-r}.
\item[RLT.] This method may be thought of as ``RF-on-RF.'' Called
  ``reinforcement learning trees'' \citep{RLT}, it constructs an
  ensemble of trees from bootstrap samples, but uses the RF
  permutation-based importance scoring method to select the most
  important variable to split each node in each tree.  After the
  ensemble is constructed, the final importance scores are obtained
  using the RF permutation scheme. The results here are produced by
  the \texttt{RLT} package \citep{RLTpack}.
\item[CTREE.] This is the ``conditional inference tree'' algorithm of
  \citet{ctree}. It follows the GUIDE approach of using significance
  tests to select a variable to split each node of a tree.  Unlike
  GUIDE, however, CTREE uses linear statistics based on a permutation
  test framework and, instead of pruning, it uses Bonferroni-type
  p-value thresholds to determine tree size.  Further, the
  significance tests employ only observations with non-missing values
  in the $X$ variable being evaluated.  Observations with missing
  values are passed through each split by means of surrogate splits as
  in CART.  Importance scores are obtained as in RFSRC, except that an
  OOB observation missing the split value at a node is randomly sent
  to the left or right child node with probabilities proportional to
  the samples sizes of the non-missing observations in the child
  nodes.

\item[CFOREST.]  This is an ensemble of CTREE trees from the
  \texttt{partykit} R package.  Instead of bootstrap samples, it takes
  random subsamples (without replacement) of about two-thirds of the
  data to construct each tree. \citet{Strobl07} showed that this
  removes a bias in RF that gives higher scores to categorical variables
  with large numbers of categories.  This is the default option in
  \texttt{partykit}, which we denote by CFOREST1.  Another option,
  which we denote by CFOREST2, is conditional permutation of the
  variables, which \citet{Strobl08} proposed for reducing the bias in
  RF towards correlated variables.
\item[LASSO.] This is linear regression with the lasso penalty. The
  importance score of an ordinal variable is the absolute value of its
  coefficient in the fitted model and that of a categorical variable
  is the average of the absolute values of the coefficients of its
  dummy variables. All variables (including dummy variables) are
  standardized to have mean 0 and variance 1 prior to model
  fitting. We use the R implementation in the \texttt{glmnet} package
  \citep{glmnet-man}.

\item[BARTM.] This is \texttt{bartMachine} \citep{Bleich14}, a
  Bayesian method of constructing a forest of regression trees using
  the BART \citep{BART} method.  The underlying model is that the
  response variable is a sum of regression tree models plus
  homoscedastic Gaussian noise. Prior distributions must be specified
  for all unknown parameters, including the set of tree structures,
  terminal node parameters, and the Gaussian noise variance.
  According to \citet{Bleich14}, the importance of a variable is given
  by the relative frequency that it appears in the splits in the
  trees.  The results here are obtained from the R package
  \texttt{bartMachine} with default parameters.
\end{description}

\section{Simulation experiments}
\label{sec:simul}

We performed 6 simulation experiments (E0--E5) involving 11 predictor
variables ($B_1$, $B_2$, $C_1$, $C_2$, $N_1$, $N_2$, $N_3$, $N_4$,
$S_1$, $S_2$, $S_3$) to compare the performance of the methods.
Variable sets $\{B_1\}$, $\{C_1\}$, $\{B_2, C_2\}$,
$\{N_1, N_2, N_3, N_4\}$, and $\{S_1, S_2, S_3\}$ are mutually
independent.  Variable $B_1$ is Bernoulli with $P(B_1 = 1) = 0.50$,
and $C_1$, $C_2$ are independent categorical variables taking values
$1,2,\ldots, 10$ with equal probability 0.10.  Variable
$B_2 = I(C_2 \leq 5)$ is a binary variable derived from
$C_2$. Variable $N_1$ is independent standard normal except in
model~E2 (see below). The triple $(N_2, N_3, N_4)$ is multivariate
normal with zero mean, unit variance, and constant correlation 0.90.
The triple $(S_1, S_2, S_3)$ is obtained by setting
$S_1 = \min(U_1, U_2)$, $S_2 = |U_1-U_2|$, and
$S_3 = 1-\max(U_1,U_2)$, where $U_1$ and $U_2$ are independent and
uniformly distributed variables on the unit interval, so that
$S_1+S_2+S_3 = 1$ and $\mbox{cor}(S_i,S_j) = -0.50$ ($i \neq j$).
Their purpose is to see if there any effects of linear dependence on the
importance scores.


Table~\ref{tab:outcome-model:new} shows the models used to generate
the dependent variable $Y = \mu(X) + \epsilon$, where $\mu(X)$ is a
function of the predictor variables and $\epsilon$ an independent
standard normal variable. Null model E0, where $Y$ is independent of
the $X$ variables, tests for bias.  The other models, which have one
or two important variables, show the effects of bias on the scores.
For each model, the scores are obtained from 1000 simulation trials,
with random samples of 400 observations in each trial.

\begin{figure}
  \centering
  \resizebox{\textwidth}{!}{\rotatebox{-90}{\includegraphics*[25,26][581,767]{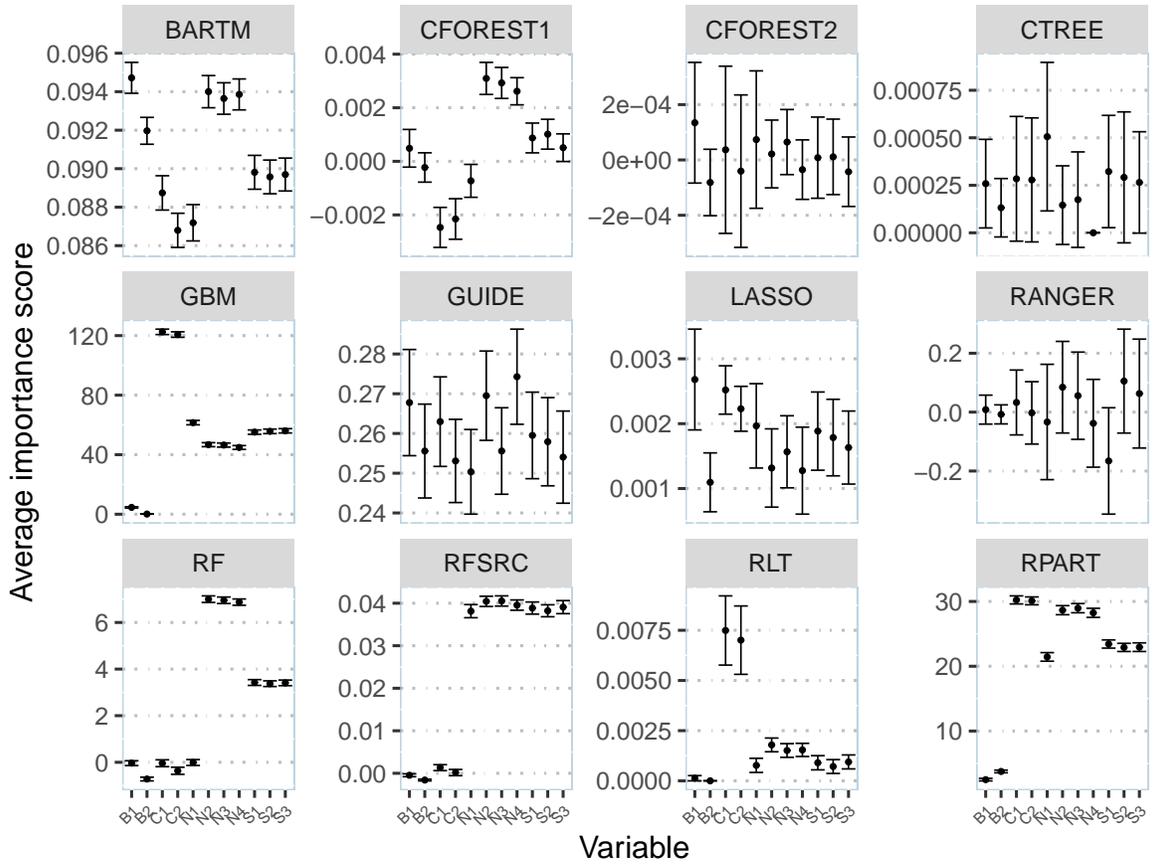}}}
  \caption{Average importance scores with 2-SE error bars for model
    E0, where predictor variables are independent of $Y$.}
  \label{fig:E0}
\end{figure}

Figure~\ref{fig:E0} shows the average scores and their 2-SE
(simulation standard error) bars for model E0. Because the 2-SE bars
should overlap if there is no selection bias, we see that only
CFOREST2, CTREE, GUIDE, and RANGER are unbiased, with RANGER and, to a
lesser degree, CFOREST2 exhibiting variance heterogeneity.  BARTM,
CFOREST1 and RF are biased towards correlated variables $N_2$, $N_3$
and $N_4$. GBM, RLT and RPART are biased towards categorical variables
$C_1$ and $C_2$. RFSRC is biased against all categorical
variables. LASSO is biased \emph{in favor} of $B_1$ but \emph{against}
$B_2$.

\begin{table}
  \centering
  \caption{Simulation models $Y = \mu(X) + \epsilon$, with $\epsilon$
    independent standard normal}
  \label{tab:outcome-model:new} \vspace{0.5em}
  \begin{tabular}{@{}ll@{}} \hline
    E0 & $\mu(X) = 0$                     \\ 
    E1 & $\mu(X) = 0.2 N_2$               \\ 
    E2 & $\mu(X) = 0.1(N_1 + N_2)$  \\     
    E3 & $\mu(X) = 0.2B_1$          \\     
    E4 & $\mu(X) = 0.2B_2 = 0.2 I(C_2 \leq 5)$ \\ 
    E5 & $\mu(X) = 0.5\{I(B_1 = 0, C_1 \leq 5) + I(B_1 = 1, C_1>5)\}$
    \\ 
    \hline
  \end{tabular}
\end{table}


\begin{figure}
  \centering
  \resizebox{\textwidth}{!}{\rotatebox{-90}{\includegraphics*[25,26][581,767]{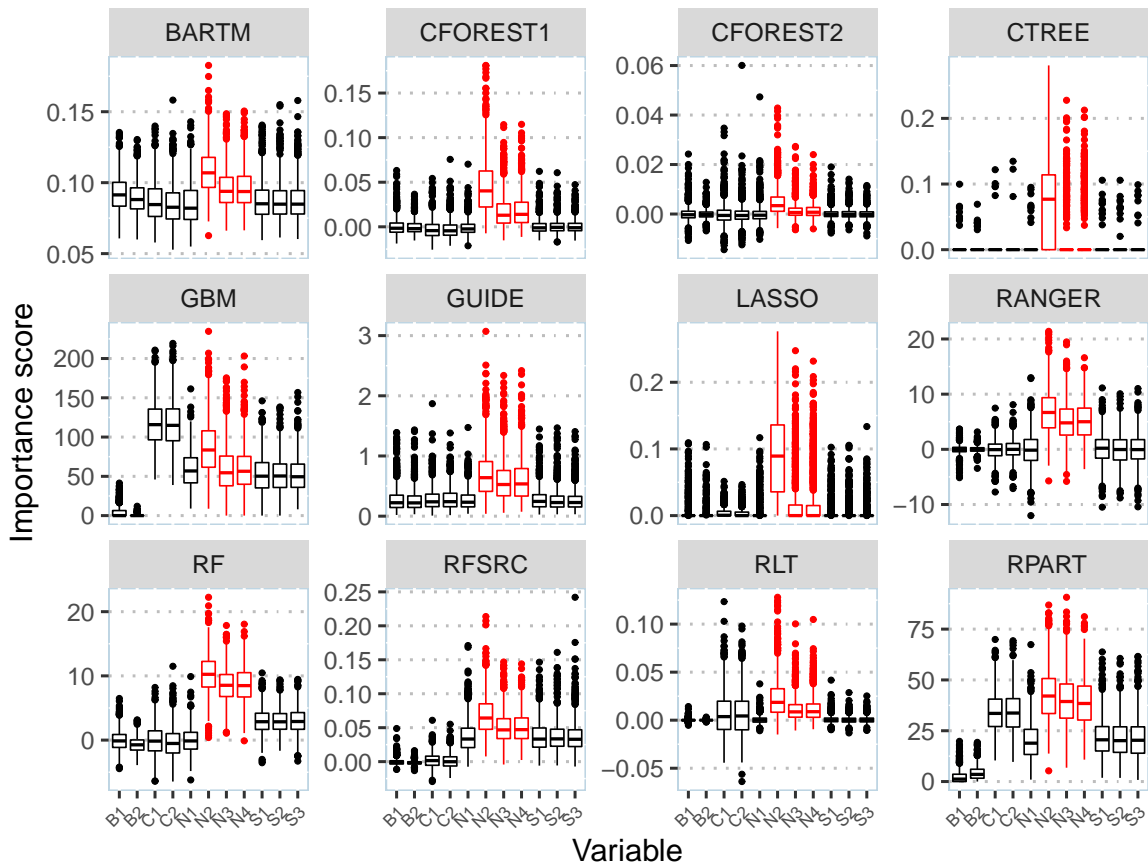}}}
  \caption{Boxplots of importance scores over 1000 trials for model
    E1, where $\mu(X) = 0.2N_2$ and $N_2, N_3$, $N_4$ are highly
    correlated.  Variables associated with $Y$ are in
    red.}  \label{fig:E1}
\end{figure}

\begin{figure}
  \centering
  \resizebox{\textwidth}{!}{\rotatebox{-90}{\includegraphics*[25,26][581,767]{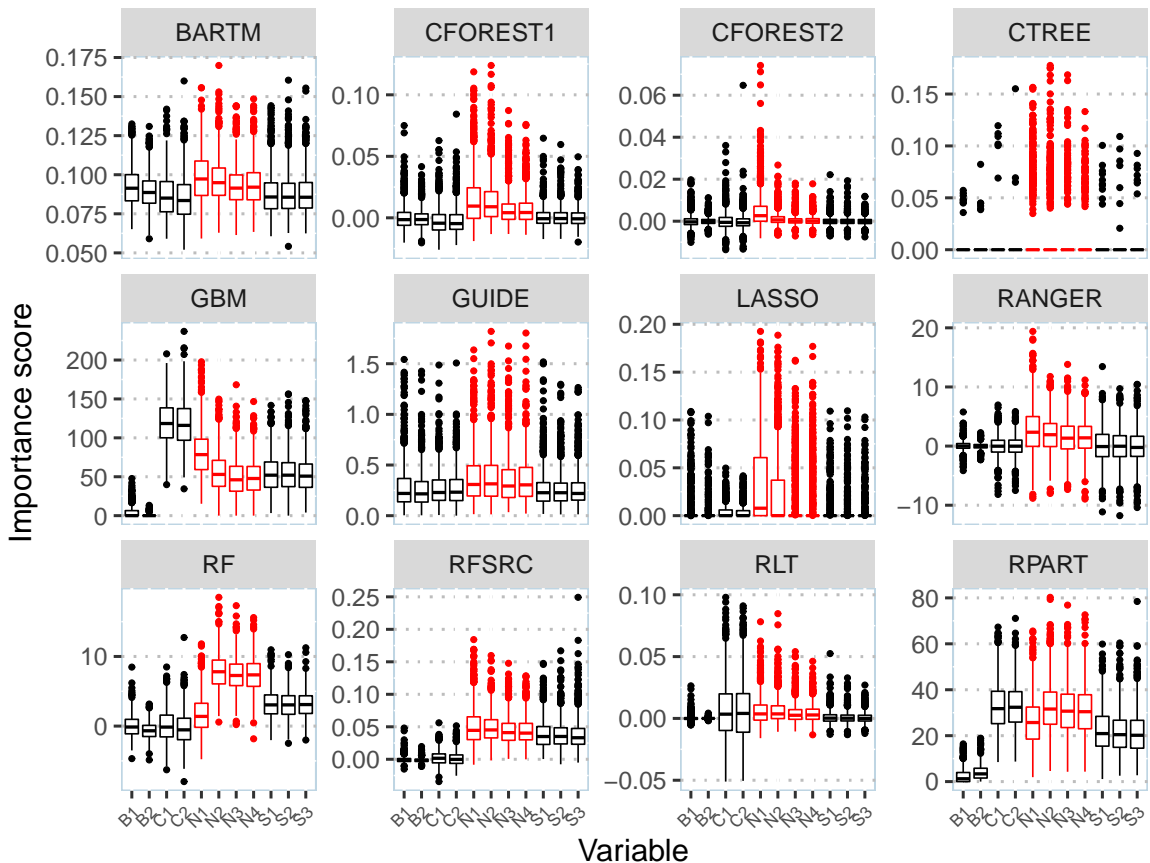}}}
  \caption{Boxplots of importance scores over 1000 trials for model E2
    where $\mu(X) = 0.1(N_1+N_2)$ and $N_2, N_3, N_4$ are highly
    correlated.  Variables with effect on $Y$ are in red.}
  \label{fig:E2}
\end{figure}

\begin{figure}
  \centering
  \resizebox{\textwidth}{!}{\rotatebox{-90}{\includegraphics*[25,26][581,767]{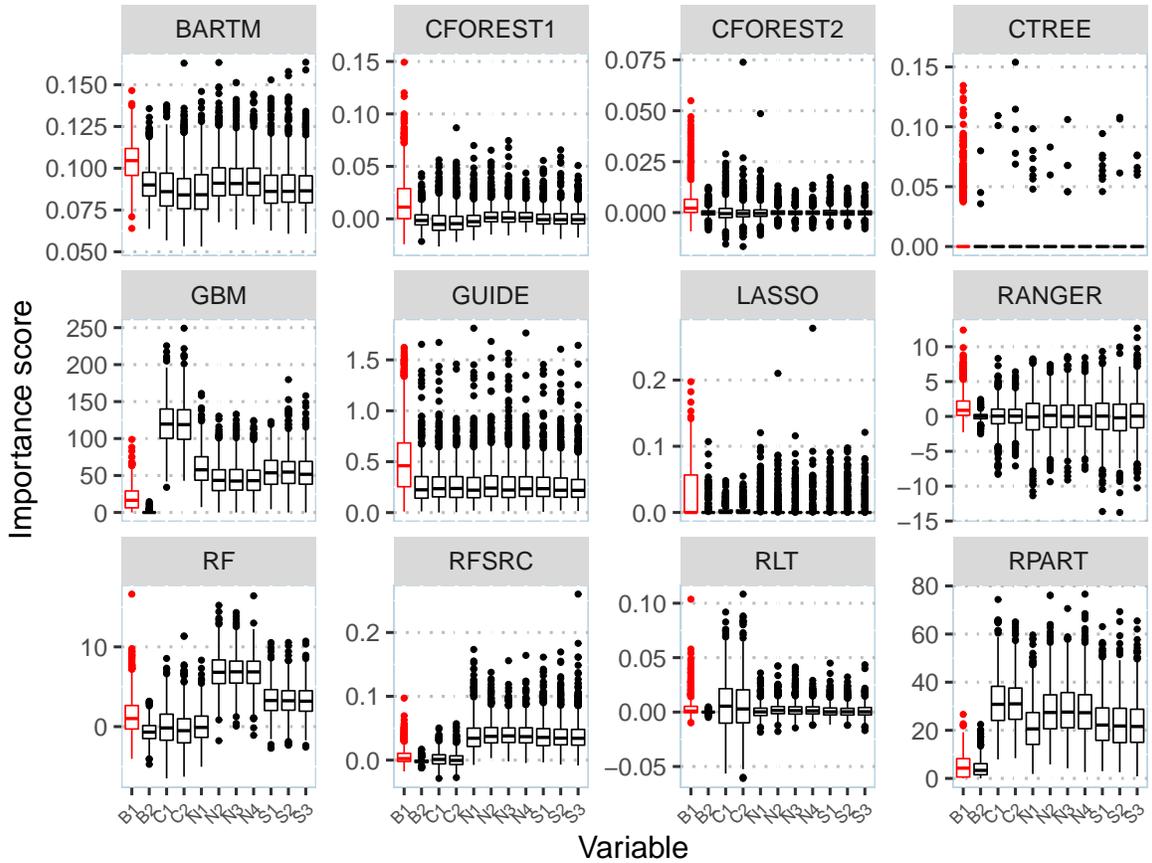}}}
  \caption{Boxplots of importance scores over 1000 trials for model
    E3, where $\mu(X) = 0.2 B_1$.  Variables with effect on $Y$ are in
    red.}
  \label{fig:E3}
\end{figure}

\begin{figure}
  \centering
  \resizebox{\textwidth}{!}{\rotatebox{-90}{\includegraphics*[25,26][581,767]{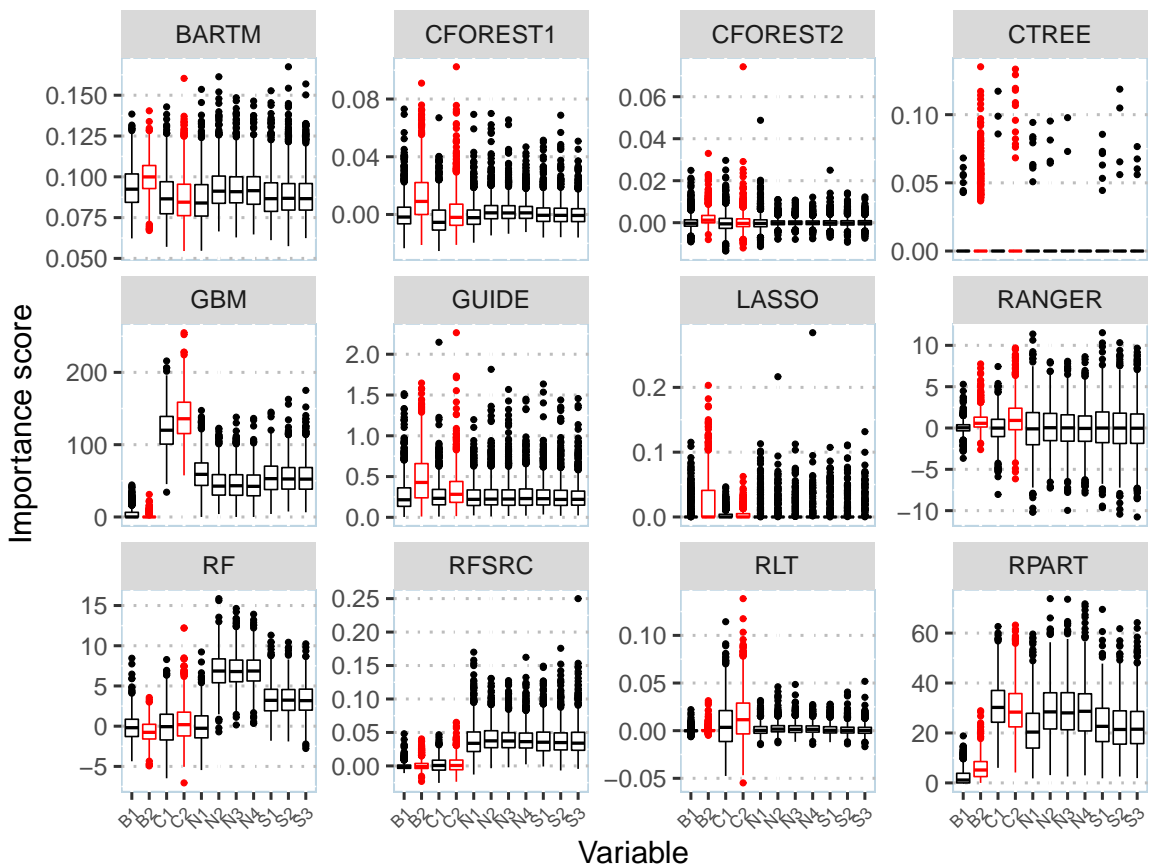}}}
  \caption{Boxplots of importance scores over 1000 trials for model
    E4, where $\mu(X) = 0.2 B_2$ and $B_2 = I(C_2 \leq 5)$.
    Variables with effect on $Y$ are in red.}
  \label{fig:E4}
\end{figure}

\begin{figure}
  \centering
  \resizebox{\textwidth}{!}{\rotatebox{-90}{\includegraphics*[25,26][581,767]{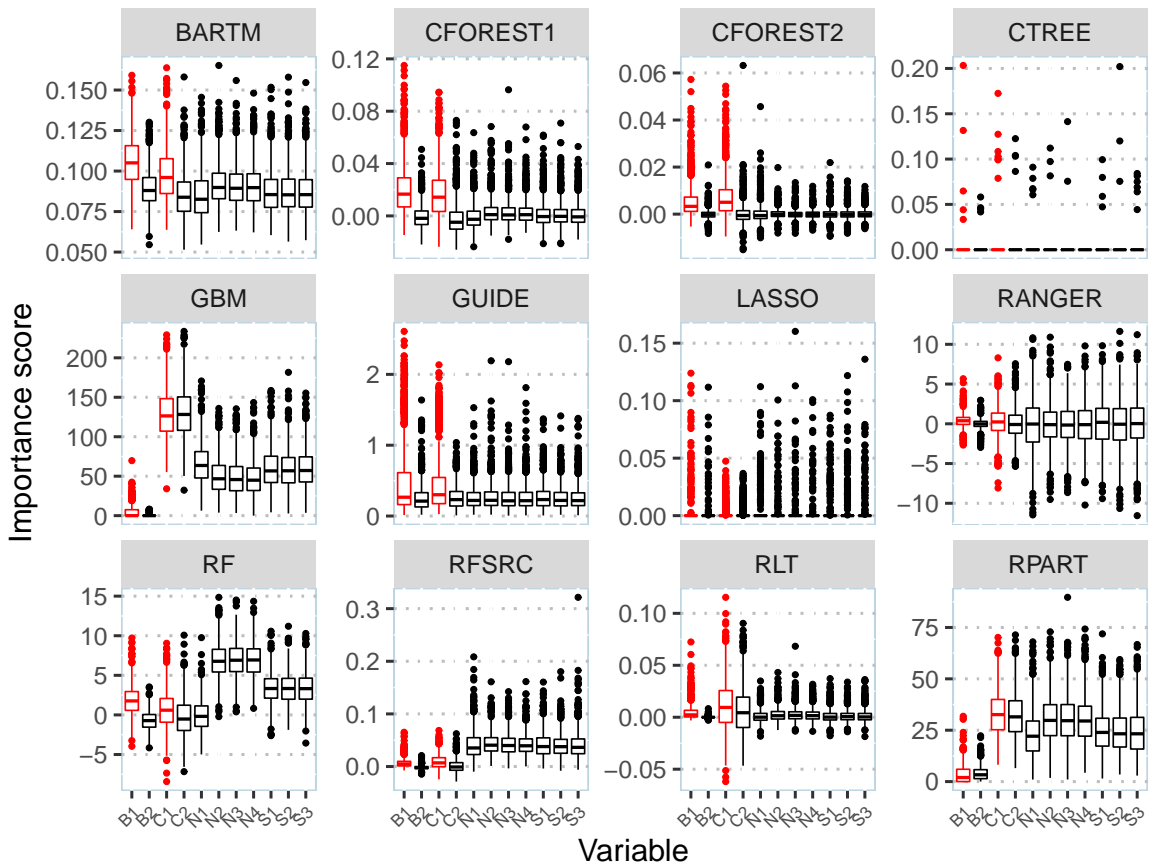}}}  
  \caption{Boxplots of importance scores over 1000 trials for model
    E5, where
    $\mu(X) = 0.5\{I(B_1 = 0, C_1 \leq 5) + I(B_1 = 1, C_1>5)\}$.
    Variables with effect on $Y$ are in red.}
  \label{fig:E5}
\end{figure}

Figures~\ref{fig:E1}--\ref{fig:E5} show boxplots of the 1000 simulated
scores for models~E1--E5. Boxplots of variables that affect $Y$ are
drawn in red. We can draw the following conclusions.
\begin{description}
\item[E1.] The model is $\mu(X) = 0.2 N_2$, but the response is also
  associated with $N_3$ and $N_4$ through their correlation with
  $N_2$.  Figure~\ref{fig:E1} shows that all but one method give their
  highest median scores to these three predictors.  The exception is
  GBM---its strong bias towards variables $C_1$ and $C_2$ makes them
  likely to be incorrectly scored higher than $N_2$, $N_3$ and $N_4$.
\item[E2.] The model is $\mu(X) = 0.1(N_1 + N_2)$, where $N_1$ is
  independent of $N_2$ but the latter is highly correlated with $N_3$
  and $N_4$. We expect their scores to be larger than those of the
  other variables, with $N_1$ and $N_2$ being roughly equal and $N_3$
  and $N_4$ close behind.  Figure~\ref{fig:E2} shows this to be true
  of all methods except GBM, RF, RLT, and RPART.  For RF and RPART,
  the presence of $N_3$ and $N_4$ raises the median score of $N_2$
  above that of $N_1$. GBM again tends to incorrectly score $C_1$ and
  $C_2$ highest. RLT also frequently incorrectly scores these two
  categorical variables higher than $N_2$, $N_3$ and $N_4$.
\item[E3.] The model is $\mu(X) = 0.2B_1$, with $B_1$ independent of
  the other predictors. All except GBM, LASSO, RF, RFSRC, RLT, and
  RPART are more likely to correctly score $B_1$ highest. GBM, RLT and
  RPART fail to do this due to bias towards $C_1$ and $C_2$.  RF fails
  due to the high correlation of $N_2$, $N_3$ and $N_4$.  CTREE and
  LASSO yield median scores of 0 for all predictors, including $B_1$.
\item[E4.] The model is $\mu(X) = 0.2B_2$ but because
  $B_2 = I(C_2 \leq 5)$, the two should have the highest median
  importance scores.  Only GUIDE, RANGER and possibly CFOREST2 have
  this property. BARTM and CFOREST1 give the highest median score to
  $B_2$ but midling median scores to $C_2$. Conversely, due to their
  bias towards categorical variables, GBM and RLT give the highest
  median score to $C_2$ but midling median scores to $B_2$. As in
  model E3, CTREE and LASSO cannot reliably identify $B_2$ or $C_2$ as
  important because both methods yield 0 median scores for all
  predictors.
\item[E5.] The model is
  $\mu(X) = 0.5\{I(B_1 = 0, C_1 \leq 5) + I(B_1 = 1, C_1>5)\}$, which
  has an interaction between $B_1$ and $C_1$.  BARTM, CFOREST1,
  CFOREST2, GUIDE, and RANGER correctly give highest median scores to
  these two predictors. GBM and RPART give $B_1$ the lowest median
  score due to their bias against binary variables.  RF incorrectly
  gives $B_1$ and $C_1$ low median scores due to its preference for
  correlated predictors. RFSRC incorrectly gives $B_1$ and $C_1$ low
  median scores due to its bias against binary and categorical
  predictors. RLT gives $C_1$ and $C_2$ the highest median scores due
  to its bias towards these two variables. CTREE and LASSO are again
  ineffective because both give zero median scores to all predictors.
\end{description}

Overall, CFOREST2, GUIDE, and RANGER are the only unbiased methods and
consequently are among the most likely to correctly identify the
important variables.

\clearpage


\section{Predictive importance}
\label{sec:predict}
``Predictive importance'' may be interpreted as the effect of a
variable on the prediction of a response, but it is not known which,
if any, of the importance scoring methods directly measures the
concept.  BARTM scores variables by their frequencies of being chosen
to split the nodes of the trees.  GBM and RPART base their scores on
decrease in impurity, and LASSO uses absolute values of regression
coefficient estimates.  CFOREST, CTREE, RANGER, RF, and RFSRC measure
change in prediction accuracy after random permutation of the
variables---an approach that \citet{Strobl08} call ``permutation
importance.''  GUIDE scores may be considered as measures of
``associative importance,'' being based on chi-squared tests of
association with the response variable at the nodes of a tree.

To see how well the scores reflect predictive importance, we need a
precise definition of the latter.  Given predictor variables
$X_1, X_2, \ldots, X_K$, consider the four models,
\begin{eqnarray}
  Y & = & \mu + \epsilon \label{eq:model0} \\
  Y &  = & f_j(X_j) + \epsilon \label{eq:model1} \\
  Y &  = & g_j(X_1, \ldots, X_{j-1}, X_{j+1}, \ldots, X_K) +
           \epsilon  \label{eq:model2} \\
  Y & = & h(X_1, X_2, \ldots, X_K) + \epsilon \label{eq:model4}
\end{eqnarray}
where $\mu$ is a constant, $f_j$, $g_j$, and $h$ are arbitrary
functions of their arguments, and $\epsilon$ is an independent
variable with zero mean and variance possibly depending on the values
of the $X$ variables. Equation~(\ref{eq:model0}) states that $E(Y)$ is
independent of the predictors, (\ref{eq:model1}) states that it
depends only on $X_j$, (\ref{eq:model2}) states that it depends on all
variables except $X_j$, and (\ref{eq:model4}) allows dependence on all
variables.  Let $\hat{\mu}$, $\hat{f}_j$, $\hat{g}_j$, and $\hat{h}$
denote estimates of $\mu$, $f_j$, $g_j$, and $h$, respectively,
obtained from a training sample and define
\begin{eqnarray*}
  S_0 & = & E (Y - \hat{\mu})^2 \\
  S_j & = & E (Y - \hat{f}_j(X_j))^2 \\
  S_{-j} & = & E (Y - \hat{g}_j(X_1,\ldots,X_{j-1},X_{j+1},\ldots,X_K))^2\\
  S & = & E (Y-\hat{h}(X_1,\ldots,X_K))^2
\end{eqnarray*}
where the expectations are computed with $\hat{\mu}$, $\hat{f}_j$,
$\hat{g}_j$, and $\hat{h}$ fixed. We call $(S_0 - S_j)$ the
\emph{marginal predictive value} of $X_j$ because it is the difference
in mean squared error between predicting $Y$ with and without $X_j$,
ignoring the other predictors.  We call $(S_{-j} - S)$ the
\emph{conditional predictive value} of $X_j$ because it is the
difference in mean squared error between predicting $Y$ without and
with $X_j$, with the other predictors included.

\begin{figure}
  \centering
  \resizebox{\textwidth}{!}{\rotatebox{-90}{\includegraphics*[180,20][439,754]{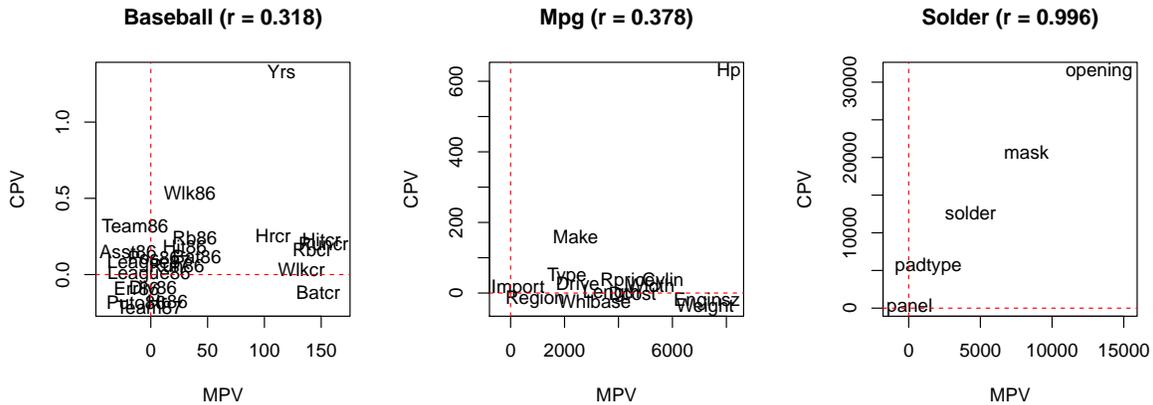}}}
  \caption{\texttt{CPV} versus \texttt{MPV} and their correlations for
    three data sets}
  \label{fig:mar-con}
\end{figure}

Correlations between the importance scores and marginal and
conditional predictive values indicate how well the former reflects
the latter.  To compute the correlations for a given data set, we need
first to estimate $\mu$, $f_j$, $g_j$, and $h$. Here we use the
average of 5 ensemble methods, namely, CFOREST, GBM, GUIDE forest,
RANGER, and RFSRC to obtain the estimates. This helps to ensure that
no scoring method has an unfair advantage.  We use leave-one-out
cross-validation to estimate $S_0$, $S_j$, $S_{-j}$, and $S$.
Specifically, given a data set $\{(y_i, x_{i1}, \ldots, x_{iK})\}$,
$i=1,2,\ldots, n$, define the vectors and matrices
\begin{eqnarray*}
  \mathbf{x}_j & = & (x_{1j},x_{2j},\ldots, x_{nj})' \\
  \mathbf{x}_j^{(-i)} & = & (x_{1j},x_{2j},\ldots, x_{i-1,j},x_{i+1,j},
                           \ldots, x_{nj})' \\
  \mathbf{X} & = & (\mathbf{x}_1,\mathbf{x}_2, \ldots, \mathbf{x}_K) \\
  \mathbf{X}^{(-i)} & = & (\mathbf{x}_1^{(-i)},\mathbf{x}_2^{(-i)}, \ldots,
                          \mathbf{x}_K^{(-i)}) \\
  \mathbf{X}_{(-j)} & = & (\mathbf{x}_1,\mathbf{x}_2, \ldots, \mathbf{x}_{j-1},
                        \mathbf{x}_{j+1}, \ldots,\mathbf{x}_K) \\
  \mathbf{X}_{(-j)}^{(-i)} & = & (\mathbf{x}_1^{(-i)},\mathbf{x}_2^{(-i)}, \ldots,
                              \mathbf{x}_{j-1}^{(-i)}, \mathbf{x}_{j+1}^{(-i)},
                              \ldots, \mathbf{x}_K^{(-i)}) 
\end{eqnarray*}
where
$(\mathbf{x}_j^{(-i)}, \mathbf{X}^{(-i)}, \mathbf{X}_{(-j)}^{(-i)})$
are $(\mathbf{x}_j, \mathbf{X}, \mathbf{X}_{(-j)})$ without the $i$th
row and $(\mathbf{X}_{(-j)}, \mathbf{X}_{(-j)}^{(-i)})$ are
$(\mathbf{X}, \mathbf{X}^{(-i)})$ without the $j$th column. Let
$(\hat{f}_j^{(-i)}, \hat{g}_j^{(-i)}, \hat{h}^{(-i)})$ denote the
function estimates of $(f_j, g_j, h)$ based on
$(\mathbf{x}_j^{(-i)}, \mathbf{X}_{(-j)}^{(-i)}, \mathbf{X}^{(-i)})$,
respectively, using the average of the 5 ensemble methods.  Let
$\bar{y} = n^{-1} \sum_k y_k$,
$\bar{y}^{(-i)} = (n-1)^{-1} \sum_{k \neq i} y_k$ and define the
leave-one-out mean squared errors 
\begin{eqnarray*}
  \hat{S}_0 & = & n^{-1} \sum_{i=1}^n (y_i - \bar{y}^{(-i)})^ 2 \\
  \hat{S}_j & = & n^{-1} \sum_{i=1}^n \{y_i - \hat{f}_j^{(-i)}(x_{ij})\}^2 \\\
  \hat{S}_{-j} & = & n^{-1} \sum_{i=1}^n \{y_i - \hat{g}_j^{(-i)}(x_{i1}, x_{i2},
                       \ldots, x_{i,j-1}, x_{i,j+1}, \ldots, x_{iK})\}^2 \\
  \hat{S} & = & n^{-1} \sum_{i=1}^n \{y_i - \hat{h}^{(-i)}(x_{i1}, x_{i2},
                      \ldots, x_{iK})\}^2.
\end{eqnarray*}
Denote the estimated marginal and conditional predictive values by
$\mathtt{MPV}_j = \hat{S}_0-\hat{S}_j$ and
$\mathtt{CPV}_j = \hat{S}_{-j}-\hat{S}$.  We compute them for the
following three real data sets.
\begin{description}
\item[Baseball.]  The data give performance and salary information of
  263 North American Major League Baseball players during the 1986
  season \citep{baseball}. The response variable is log-salary and
  there are 22 predictor variables; see \citet{HV95} and references
  therein for definitions of the variables. The plot on the left side
  of Figure~\ref{fig:mar-con} shows a rather weak correlation of 0.318
  between $\mathtt{CPV}$ and $\mathtt{MPV}$.  Variable \texttt{Yrs}
  (number of years in the major leagues) has high values of
  $\mathtt{MPV}$ and $\mathtt{CPV}$ but \texttt{Batcr} (number of
  times at bat during career) has a high value of $\mathtt{MPV}$ and a
  negative value of $\mathtt{CPV}$. This implies that \texttt{Batcr}
  is an excellent predictor if it is used alone, but its addition
  after the other variables are included does not increase accuracy.

\item[Mpg.] This data set gives the characteristics, price, and dealer
  cost of 428 new model year 2004 cars and trucks \citep{mpg}. We use
  14 variables to predict city miles per gallon (mpg). The middle
  panel of Figure~\ref{fig:mar-con} shows that \texttt{Hp}
  (horsepower) has the highest values of $\mathtt{MPV}$ and
  $\mathtt{CPV}$. Variable \texttt{Make} (which has 38 categorical
  values) has the second highest $\mathtt{CPV}$ but its $\mathtt{MPV}$
  is below average, indicating that its predictive power is mainly
  derived from interactions with other variables. The correlation
  between $\mathtt{CPV}$ and $\mathtt{MPV}$ is 0.378.

\item[Solder.]  \citet{CH92} used the data from a circuit board
  soldering experiment to demonstrate Poisson regression in R. The
  data, named \texttt{solder.balance} in the \texttt{rpart} R package,
  give the number of solder skips in a 5-factor unreplicated
  $3 \times 2 \times 4 \times 10 \times 3$ factorial experiment.
  Because not all scoring methods are applicable to Poisson
  regression, we use least squares with dependent variable the square
  root of the number of solder skips.
  The right panel of Figure~\ref{fig:mar-con} shows that
  $\mathtt{CPV}$ and $\mathtt{MPV}$ are almost perfectly
  correlated. This is a consequence of the factorial design.
\end{description}

\begin{table}
  \centering
  \caption{Correlations between importance scores \texttt{VI} and
    marginal and conditional predictive values \texttt{MPV} and
    \texttt{CPV}}
  \label{tab:comp-cor} \vspace{0.5em}
  \begin{tabular}{l|ll|ll|ll}
    & \multicolumn{2}{c}{Baseball} & \multicolumn{2}{c}{Mpg} 
    & \multicolumn{2}{c}{Solder} \\
      Method & \texttt{MPV} & \texttt{CPV} 
             & \texttt{MPV} & \texttt{CPV} 
             & \texttt{MPV} & \texttt{CPV} \\ \hline
    BARTM    & 0.75 & 0.68 & 0.85 &  0.48 &  0.4  &   0.46 \\
    CFOREST1 & 0.87 & 0.1  & 0.82 &  0.28 &  1    &   1    \\
    CFOREST2 & 0.82 & 0.16 & 0.69 &  0.78 &  0.99 &   1    \\
    CTREE    & 0.4  & 0.07 & 0.65 &  0.54 &  0.99 &   1    \\
    GBM      & 0.8  & 0.14 & 0.62 &  0.88 &  0.99 &   0.98 \\
    GUIDE    & 0.99 & 0.3  & 0.94 &  0.24 &  0.9  &   0.92 \\
    LASSO    & 0.19 & 0.59 & 0.75 &  0.55 &  0.73 &   0.76 \\
    RANGER   & 0.97 & 0.18 & 0.96 &  0.33 &  1    &   1    \\
    RF       & 0.83 & 0.16 & 0.54 &  0.28 &  0.87 &   0.91 \\
    RFSRC    & 0.79 & 0.02 & 0.72 &  0.8  &  1    &   1    \\
    RLT      & 0.69 & 0    & 0.67 &  0.77 &  0.99 &   1    \\
    RPART    & 0.92 & 0.2  & 0.85 &  0.44 &  0.9  &   0.93 \\
    \hline
  \end{tabular}
\end{table}

\begin{figure}
  \centering
  \resizebox{\textwidth}{!}{\rotatebox{-90}{\includegraphics*[170,16][447,777]{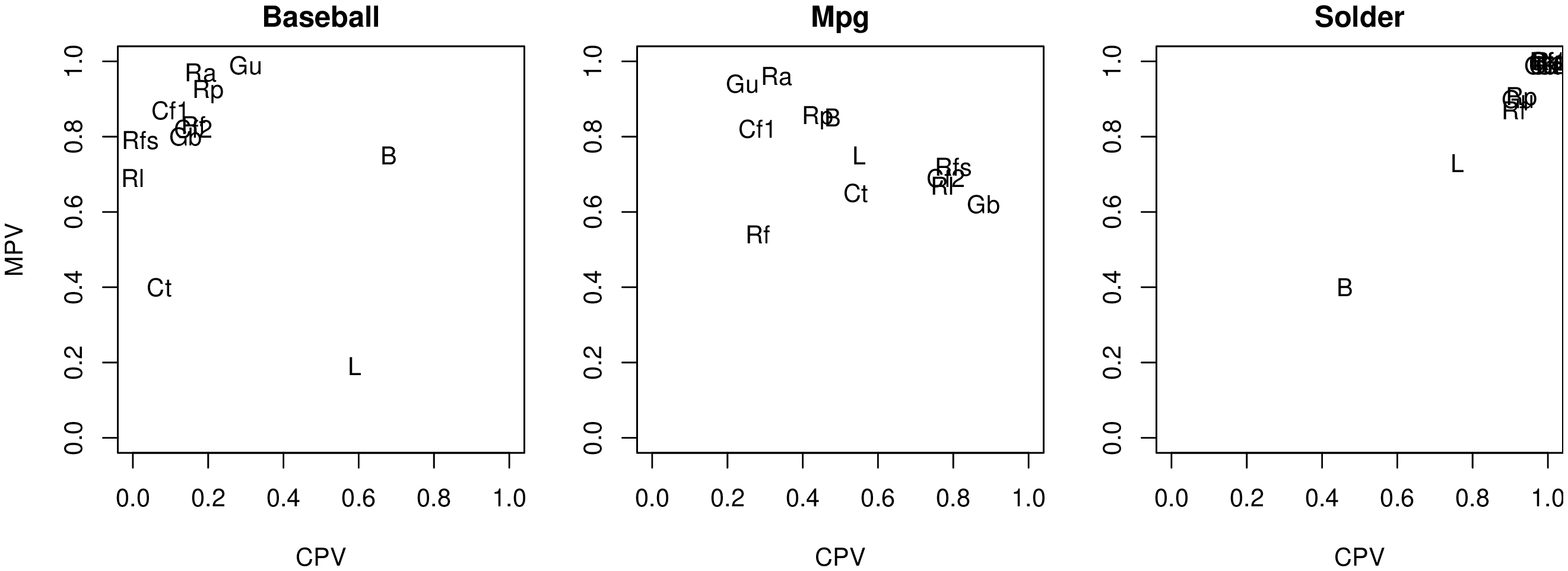}}}
  \caption{Plots of (cor(\texttt{VI}, \texttt{CPV}), cor(\texttt{VI},
    \texttt{MPV})) for three data sets; \texttt{B} = BARTM, \texttt{Cf1} = CFOREST1,
    \texttt{Cf2} = CFOREST2, \texttt{Ct} = CTREE, \texttt{Gb} = GBM,
    \texttt{Gu} = GUIDE, \texttt{L} = LASSO, \texttt{Ra} = RANGER,
    \texttt{Rf} = RF, \texttt{Rfs} = RFSRC, \texttt{Rl} = RLT,
    \texttt{Rp} = RPART}
  \label{fig:mpv-cpv}
\end{figure}

Table~\ref{tab:comp-cor} gives the correlations between the importance
scores \texttt{VI} and each of \texttt{MPV} and \texttt{CPV} for each
method and Figure~\ref{fig:mpv-cpv} shows them graphically. The
results may be summarized as follows.
\begin{description}
\item[Baseball.]  The importance scores are highly correlated with
  \texttt{MPV} for GUIDE and RANGER, but not for LASSO where there is
  barely any correlation.  On the other hand, the scores are weakly
  correlated with \texttt{CPV} for all methods except BARTM and LASSO.
\item[Mpg.] GUIDE and RANGER are again the two methods with importance
  scores most highly correlated with \texttt{MPV}; the correlations for
  the other methods range from 0.54 for RF to 0.85 for BARTM and
  RPART. For \texttt{CPV}, GBM has the highest correlation of 0.88,
  followed by RFSRC (0.0.80) and CFOREST2 (0.78).
\item[Solder.] Owing to the almost perfect correlation between
  \texttt{MPV} and \texttt{CPV}, their correlations with the
  importance scores are almost the same. BARTM and LASSO are the only
  two methods with correlations substantially below 0.90, indicating
  that they are measuring something besides \texttt{MPV} and
  \texttt{CPV}.
\end{description}
Across the three data sets, the importance scores of all methods,
except for \texttt{BARTM} and \texttt{LASSO}, are consistent with
\texttt{MPV}, with GUIDE, RANGER and RPART showing the highest
consistency. Consistency with \texttt{CPV} is weaker and more variable
between data sets.

\section{Thresholding}
\label{sec:threshold}
It is useful to have a score threshold to identify the variables that
are independent of the response. This is particularly desirable if the
number of variables is large.  Of the 12 scoring methods, only BARTM
and GUIDE currently provide thresholds.  We call a variable
``unimportant'' if it is independent of the response variable and
``important'' otherwise.  Under the null hypothesis $H_0$ that all
variables are unimportant, we define a ``Type I error'' as that of
declaring at least one to be important.  To control the probability of
this error at significance level $\alpha$, \citet{Bleich14} randomly
permute the $Y$ values several times, keeping the $X$ values
fixed. They construct a BARTM forest to each set of permuted data,
derive several candidate thresholds from the permutation distributions
of the variable selection frequencies, and use cross-validation to
choose among them. 

GUIDE similarly permutes the $Y$ values, keeping the $X$ values fixed.
For $j=1,2,\ldots, 300$, let $u_j = \max_i \mathtt{VI}(X_i)$ denote
the maximum value of the GUIDE importance scores for the $j$th
permuted data set and let $u^*(\alpha)$ be the $(1-\alpha)$-quantile
of the distribution of $\{u_1, u_2, \ldots, u_{300}\}$.  Under $H_0$,
the probability that one or more importance scores exceeds the value
of $u^*(\alpha)$ is approximately $\alpha$.

Bias adjustment of the importance scores defined in
equation~(\ref{eq:guide:scaled}) requires one level of permutation.
Calculation of $u^*(\alpha)$ requires a second level of
permutation. To skip the second level, GUIDE uses the following
approximation. In the permutations for bias adjustment, let
$v_b = \max_i v^*_b(X_i)$, $b=1,2,\ldots,B$, denote the maximum
unadjusted score, where $v^*_b(X_i)$ is defined above
equation~(\ref{eq:guide:scaled}).  Let $v^*(\alpha)$ denote the
$(1-\alpha)$-quantile of $\{v_1, v_2, \ldots, v_B\}$. Let $s(X_i)$ be
the unadjusted score for the unpermuted (real) data defined in
(\ref{eq:guide:raw}). Finally, let $m$ denote the number of values of
$s(X_i)$ greater than $v^*(\alpha)$. We declare the variables with the
top $m$ values of the bias-adjusted scores $\mathtt{VI}(X_i)$ to be
important. Let $\tilde{v}(\alpha)$ denote the average of the $m$th and
$(m+1)$th largest values of $\mathtt{VI}(X_i)$.  The GUIDE normalized
importance scores are $\mathtt{VI}(X_i)/\tilde{v}(\alpha)$, so that
variables with normalized scores less than 1.0 are considered
unimportant.

\begin{table}
  \centering
  \caption{Important variables (in alphabetical order
    for BARTM, in decreasing importance for GUIDE) for $\alpha = 0.05$}
  \label{tab:thresh} \vspace{1em}
  {\RaggedRight
    \begin{tabular}{l|p{2.5in}|p{2.5in}}
      Data & \multicolumn{1}{c|}{BARTM} & \multicolumn{1}{c}{GUIDE} \\ \hline
      COVID & \texttt{diabetes}, \texttt{race=Black} \texttt{or} \texttt{African
              American}, \texttt{race=Unknown}, \texttt{race=White} 
                                        & \texttt{renal}, \texttt{charlson},
                                          \texttt{agecat}, \texttt{MI}, \texttt{CHF},
                                          \texttt{dementia}, \texttt{PVD},
                                          \texttt{cerebro}, \texttt{cancer},
                                          \texttt{diabetes}, \texttt{race},
                                          \texttt{CPD}, \texttt{sex},
                                          \texttt{metastatic}, \texttt{hemipara},
                                          \texttt{modsevliv}, \texttt{mildliver} \\ \hline
      Baseball &  \texttt{Hitcr}, \texttt{Rbcr}, 
                 \texttt{Runcr}, \texttt{Yrs} 
                                        & \texttt{Batcr}, \texttt{Hitcr}, \texttt{Runcr}, \texttt{Rbcr},
                                          \texttt{Wlkcr}, \texttt{Yrs}, \texttt{Hrcr}, \texttt{Hit86},
                                          \texttt{Rb86}, \texttt{Bat86}, \texttt{Wlk86}, \texttt{Run86},
                                          \texttt{Hr86}, \texttt{Pos86}, \texttt{Puto86} \\ \hline
      Mpg & \texttt{Cylin=3}, \texttt{Cylin=4}, \texttt{Enginsz}, 
            \texttt{Hp},  \texttt{Make=Honda}, \texttt{Make=Kia},
            \texttt{Make=Toyota}, \texttt{Type=car}, \texttt{Weight} 
                                        & \texttt{Weight}, \texttt{Enginsz}, \texttt{Cylin}, \texttt{Hp},
                                          \texttt{Dcost}, \texttt{Rprice}, \texttt{Width}, \texttt{Whlbase},
                                          \texttt{Drive}, \texttt{Type}, \texttt{Make}, \texttt{Length},
                                          \texttt{Region} \\ \hline
      Solder & \texttt{mask=B6}, \texttt{opening=small} 
                                        & \texttt{opening}, \texttt{mask}, \texttt{solder}, \texttt{padtype} \\
      \hline
    \end{tabular}
  }
\end{table}

Table~\ref{tab:thresh} lists the variables found to be important by
BARTM and GUIDE in the COVID, Baseball, Mpg, and Solder data sets,
using $\alpha = 0.05$. GUIDE orders the important variables by their
\texttt{VI} values, but BARTM does not order them. The table shows
that BARTM tends to find fewer important variables than
GUIDE. Besides, because it transforms each categorical variable into
several indicator variables, BARTM may find some indicators important
and other indicators unimportant. For example in the SOLDER data,
BARTM finds only level \texttt{B6} of \texttt{mask} and level
\texttt{small} of \texttt{opening} important.

\section{Missing values}
\label{sec:miss}
Among the 12 scoring methods, only CFOREST1, GUIDE, RPART, and RFSRC
are directly applicable to data with missing values.  By treating
missing values as a special type of observation in GUIDE as described
in Section~\ref{sec:guide}, its importance scores remain unbiased when
there are missing values.  To demonstrate this as well as observe the
effect of missing values on CFOREST1, RPART and RFSRC, we apply the
methods to a data set from a Bureau of Labor Statistics Consumer
Expenditure (CE) Survey.  The data consist of answers to more than 400
questions from 6464 survey respondents. The dependent variable is
\texttt{INTRDVX}, the amount of interest and dividends from the
previous year. About 25\% of the values of \texttt{INTRDVX} are
missing, due to the question being inapplicable or the respondent
refusing to answer it. Here we use the 4693 respondents with
non-missing \texttt{INTRDVX} to obtain importance scores for its
prediction. About 20\% of the other variables have missing values,
with 67 of them having more than 95\% missing, including
\texttt{STOCKX} (value of directly-held stocks, bonds, mutual funds,
etc.), which may be expected to be a good predictor of
\texttt{INTRDVX}. See \citet{lecl19,LZZZ20} for more information on
the variables.

\begin{figure}
  \centering
  \resizebox{\textwidth}{!}{\rotatebox{-90}{\includegraphics*[21,26][300,768]{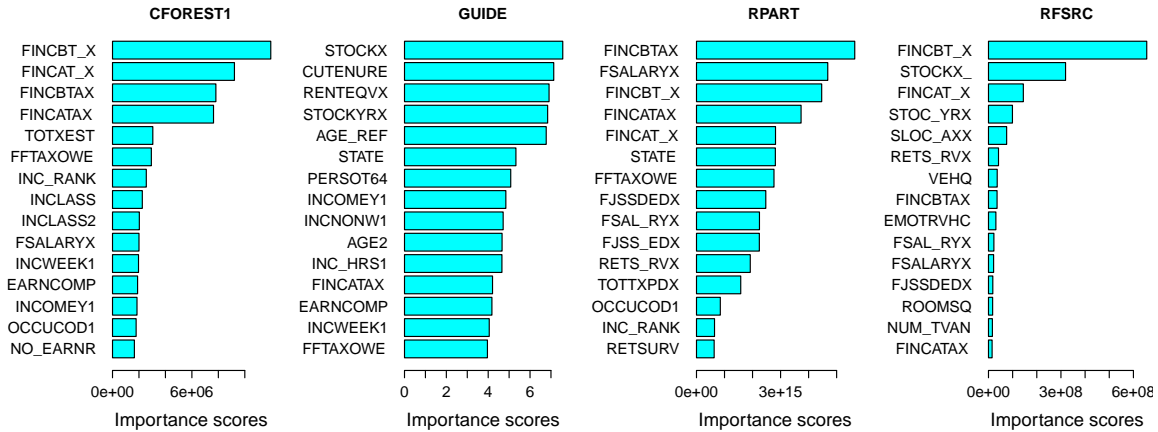}}}
  \caption{Variables with 15 highest importance scores in CE data}
  \label{fig:bls:impscr}
\end{figure}

Figure~\ref{fig:bls:impscr} shows barplots of the scores of the top 15
variables for each method. \texttt{STOCKX} is ranked most important by
GUIDE and second most important by RFSRC, but it is not ranked in the
top 15 by CFOREST1 and RPART.  At least one of \texttt{FINCBTAX}
(income before tax) or \texttt{FINCATAX} (income after tax) is in the
top 15 of all four methods. These two variables have no missing
values.


\begin{figure}
  \centering
  \resizebox{\textwidth}{!}{\rotatebox{-90}{\includegraphics*[32,17][580,753]{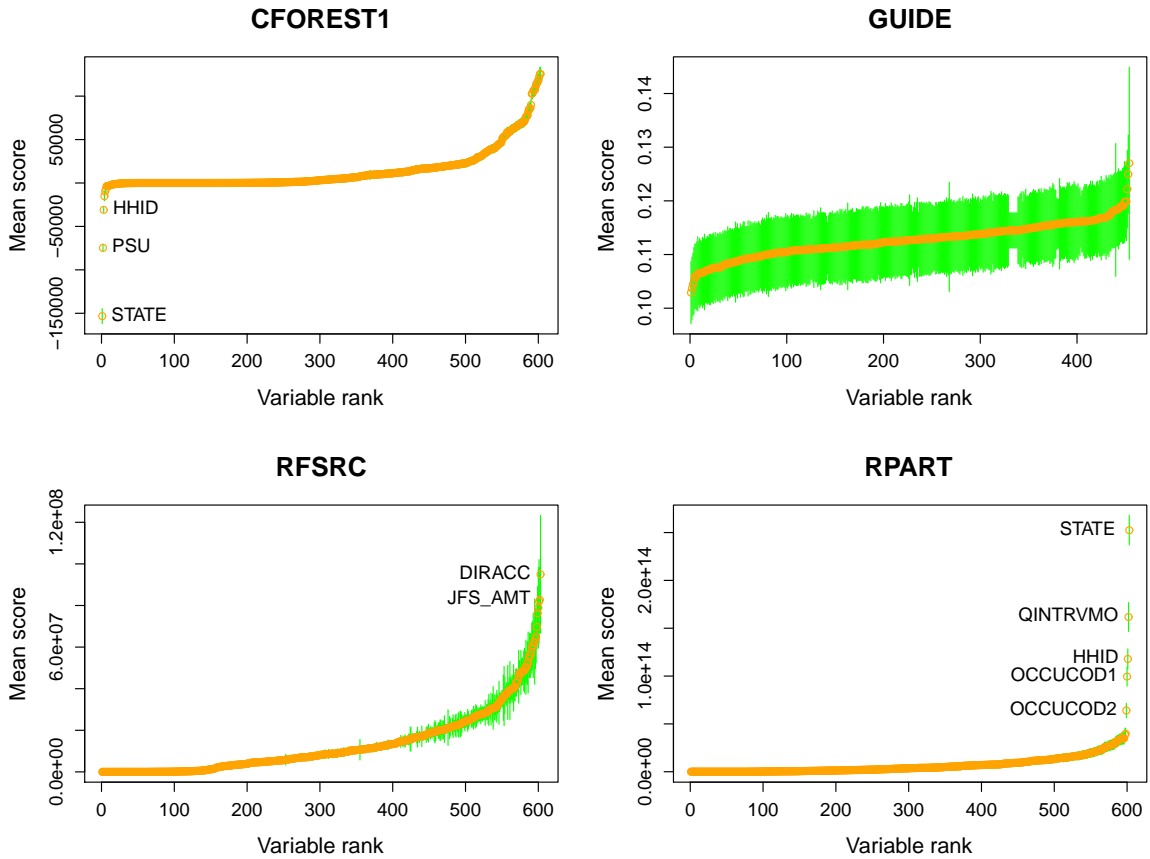}}}
  \caption{Mean importance scores $\overline{\mathtt{VI}}$ (orange)
    and 2-SE bars (green) from 1000 random permutations of the
    response variable for CE data. Variables ordered by increasing
    mean scores.  GUIDE has fewer variables because it combines
    missing-value flag variables with their associated variables.}
  \label{fig:ce:boxplots}
\end{figure}

We can use the same procedure that produced
Figure~\ref{fig:covid:perm} to find out if there is bias in the
importance scores by randomly permuting the \texttt{INTRDVX} values
while holding the values of the predictor variables fixed.  Let $J$ be
the number of permutations and $m_j(k)$ be the importance score of
variable $X_k$ in permutation $j$ ($j = 1, 2, \ldots, J$).
Figure~\ref{fig:ce:boxplots} plots $\bar{m}(k) = J^{-1} \sum_j m_j(k)$
(in orange, arranged in increasing order) and their 2-SE error bars
(in green) for each method, with $J=1000$.  GUIDE is the only method
with unbiased scores as its 2-SE bars completely overlap.  The other
three methods are biased.  CFOREST1 is particularly biased against the
three high-level categorical variables \texttt{HHID} (household
identifier, 46 levels), \texttt{PSU}, (primary sampling unit, 21
levels), and \texttt{STATE} (39 levels).  RPART is biased in favor of
\texttt{STATE}, \texttt{HHID}, and two 15-level categorical variables
\texttt{OCCUCOD1} (respondent occupation) and \texttt{OCCUCOD2}
(spouse occupation).  RFSRC is biased towards the binary variable
\texttt{DIRACC} (access to living quarters) and the continuous
variable \texttt{JFS\_AMT} (annual value of food stamps).

\section{Conclusion}
\label{sec:conclusion}
We have presented an importance scoring method based on the GUIDE
algorithm and compared it with 11 other methods in terms of bias and
consistency with two measures of predictive importance.  We say that a
method is unbiased if the expected values of its scores are equal when
all variables are independent of the response variable.  We find that
if the data do not have missing values, only CFOREST2, CTREE, GUIDE,
and RANGER are unbiased.  RF and RFSRC are biased against categorical
variables; GBM is biased towards high-level categorical variables and
against binary variables; RPART is biased against binary variables;
and RLT is biased towards high-level categorical variables and against
binary variables.  BARTM, CFOREST1 and LASSO have biases that are hard
to characterize.  Only CFOREST1, GUIDE, RPART, and RFSRC are directly
applicable to data with missing values. Among them, only GUIDE is
unbiased.  Unbiasedness of GUIDE is built into the method, through
bias correction by random permutation of the values of the response
variable.  The technique is applicable to any scoring method that is
not extremely biased, albeit at the cost of increasing computational
time several hundred fold.

\begin{figure}
  \centering
  \resizebox{\textwidth}{!}{\rotatebox{-90}{\includegraphics*[26,20][558,742]{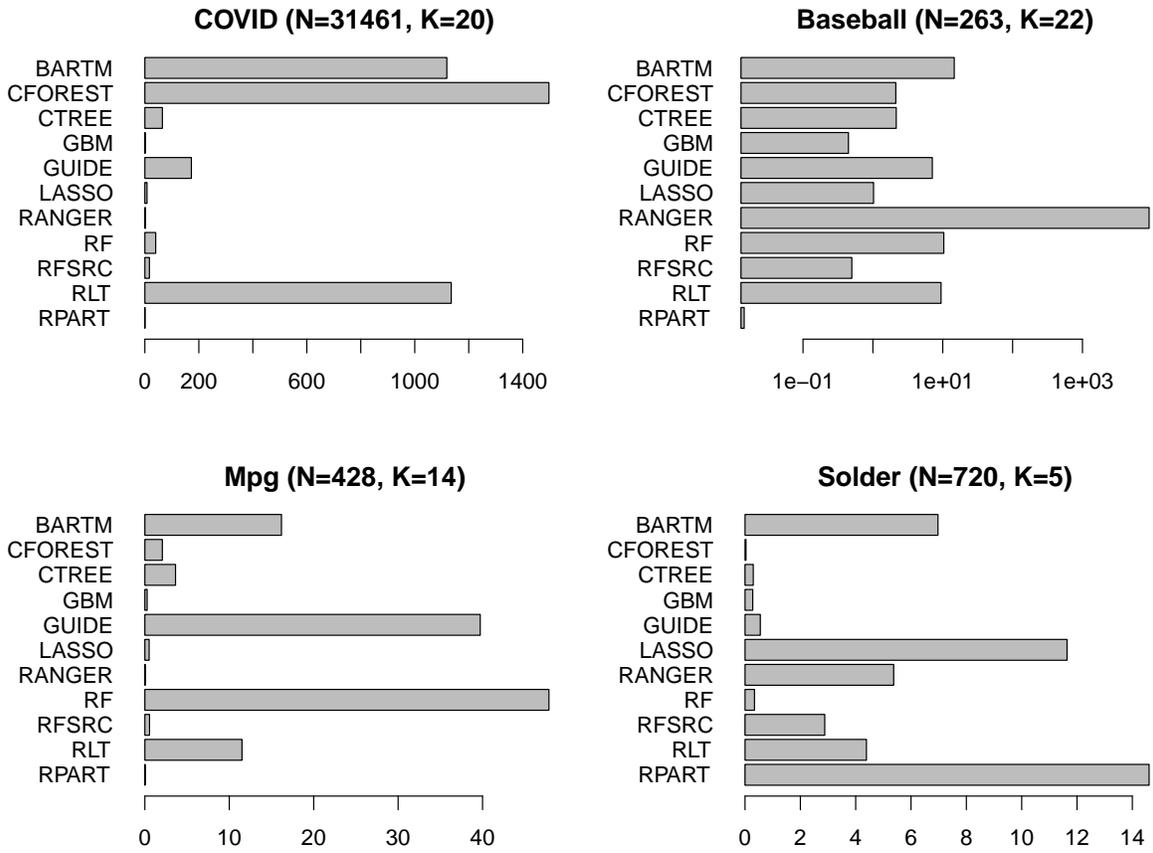}}}
  \caption{Average CPU times (sec.) for one set of importance scores;
    N is the sample size and K is the number of variables}
  \label{fig:cpu}
\end{figure}

Figure~\ref{fig:cpu} shows average computation times in seconds for
each method to calculate one set of importance scores for the four
data sets without missing values.  The computations were performed on
an Intel Xeon 2.40GHz computer with 56 cores and 240 GB memory. The
timings are averages of 3 replications, to reduce the variability of
randomized methods (CFOREST, GBM, LASSO, RANGER, RF, RFSRC, RLT) that
employ random number seeds.  In real applications the randomized
methods will take much longer, because their importance scores have to
be averaged over many replications.  The barplot for the baseball data
is drawn on a log scale due to the unusually long computation time for
RANGER; we attribute the reason to there being 3 categorical variables
each with 23 levels in that data.

We use three data sets to examine whether the importance scores
correlate well with two measures of predictive power, namely marginal
predictive value (where other variables are ignored) and conditional
predictive value (where other variables are fitted first).  We find
that the scores of many methods are highly correlated ($> 0.80$) with
marginal predictive value, the exceptions being BARTM, CTREE, and
LASSO.  Correlations with conditional predictive values, however, are
generally low, except for CFOREST2, GBM, RFSRC, and RLT, where the
correlations range from 0.77 to 0.88 in one data set.

Finally, we show how GUIDE constructs $100(1-\alpha)$\% threshold
scores for distinguishing important from unimportant variables. The
thresholds are constructed such that if all predictors are independent
of the response, the probability that one or more of them score above
the thresholds is $\alpha$.  As with bias correction, the GUIDE
threshold technique is applicable to other methods.



\vskip 0.2in
\bibliography{paper}

\end{document}